\documentclass[conference,a4paper]{APSIPA2025}
\usepackage{amsmath}
\usepackage{graphicx}
\usepackage{multirow}
\usepackage{threeparttable}
\usepackage{comment}
\usepackage{subcaption}
\usepackage[backend=biber,style=ieee,]{biblatex}
\AtBeginBibliography{\footnotesize}
\usepackage{geometry}
\usepackage{fancyhdr}

\fancypagestyle{firststyle}{
  \fancyhf{}
  \fancyhead[C]{2025 Asia Pacific Signal and Information Processing Association Annual Summit and Conference (APSIPA ASC)}
}

\begin{document}

\title{HyTver: A Novel Loss Function for Longitudinal Multiple Sclerosis Lesion Segmentation.}

\author{
\authorblockN{
Dayan Perera \authorrefmark{1}, 
Ting Fung Fung\authorrefmark{1} and 
Vishnu Monn Baskaran\authorrefmark{1}
}

\authorblockA{
\authorrefmark{1}
School of Information Technology, Monash University, Malaysia Campus \\
E-mail: \{dayan.perera, ting.fung, vishnu.monn\}@monash.edu
}


}

\maketitle
\thispagestyle{firststyle}
\pagestyle{fancy}

\begin{abstract}
    Longitudinal Multiple Sclerosis Lesion Segmentation is a particularly challenging problem that involves both input and output imbalance in the data and segmentation. Therefore in order to develop models that are practical, one of the solutions is to develop better loss functions. Most models naively use either Dice loss or Cross-Entropy loss or their combination without too much consideration. However, one must select an appropriate loss function as the imbalance can be mitigated by selecting a proper loss function. In order to solve the imbalance problem, multiple loss functions were proposed that claimed to solve it. They come with problems of their own which include being too computationally complex due to hyperparameters as exponents or having detrimental performance in metrics other than region-based ones. We propose a novel hybrid loss called HyTver that achieves good segmentation performance while maintaining performance in other metrics. We achieve a Dice score of 0.659 while also ensuring that the distance-based metrics are comparable to other popular functions. In addition, we also evaluate the stability of the loss functions when used on a pre-trained model and perform extensive comparisons with other popular loss functions.
    
\end{abstract}

\section{Introduction}

Multiple Sclerosis (MS) is an autoimmune disease that affects the Central Nervous System (CNS). The disease affects more than 2.5 million people worldwide and is characterized by lesions in the CNS, which can cause severe physical and cognitive impairments or neurological defects among young adults \cite{dilokthornsakul_valuck_nair_corboy_allen_campbell_2016}. Due to the lack of a complete cure, monitoring and careful control of symptoms is the only solution for patients. Manual delineation of the lesion is also time consuming; therefore, physicians increasingly rely on medical imaging and deep learning technologies for diagnosis and monitoring\cite{ghasemi_razavi_nikzad_2017}. 

\begin{figure}
\centering
    \begin{subfigure}{.25\textwidth}
        \centering
        \includegraphics[scale=0.125]{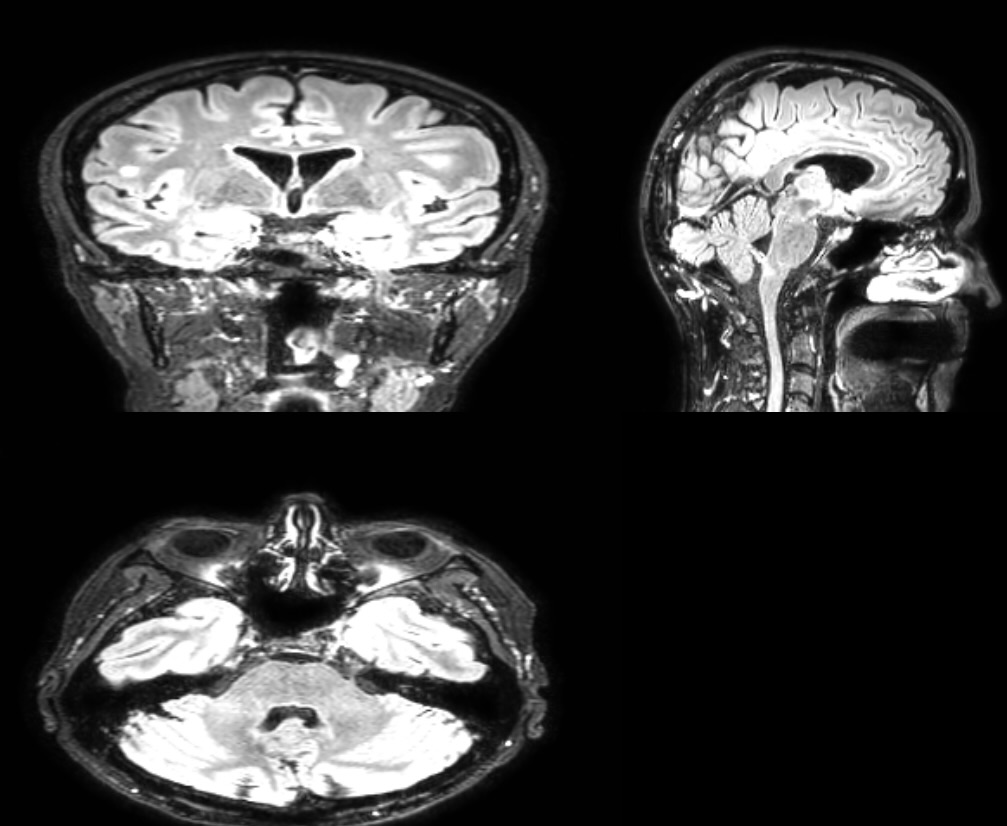}
        \caption{baseline scan}
        \label{fig:sub1}
    \end{subfigure}%
    \begin{subfigure}{.25\textwidth}
        \centering
        \includegraphics[scale=0.125]{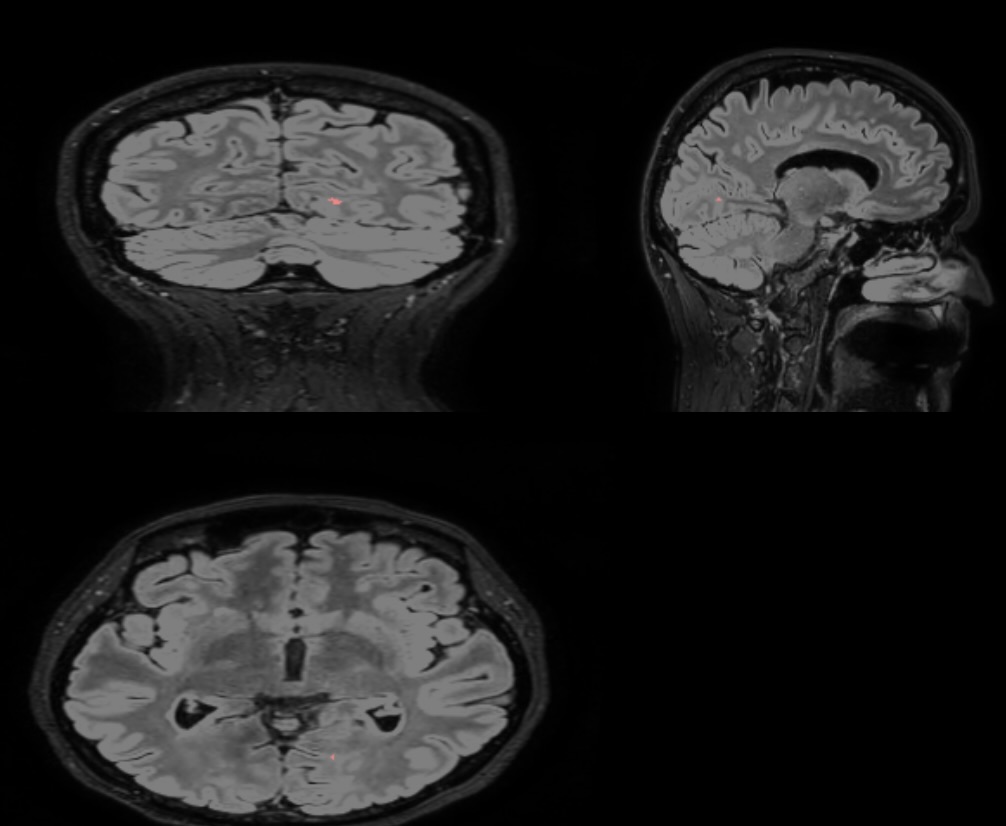}
        \caption{Follow-up scan}
        \label{fig:sub2}
    \end{subfigure}
    \caption{Baseline and Follow-up scans of the MSSEG dataset}
    \label{fig:1}
\end{figure}

One of the most widely used deep learning architectures is the U-Net \cite{weissleder_nahrendorf_2015} introduced in 2015. Its variants, such as the V-Net \cite{milletari_navab_ahmadi_2016}, are also gaining popularity. Notable performance in MS lesion segmentation was achieved by the method introduced by \cite{valverde_cabezas_roura_gonzález-villà_pareto_vilanova_ramió-torrentà_rovira_oliver_lladó_2017} in 2017 reporting a dice score of 87.2. Therefore, the focus shifted to achieving more detailed disease monitoring using deep learning. The next step which is longitudinal MS segmentation, which uses scans from two or more time points to identify newly emerging lesions. These new lesions are crucial markers of disease progression and inform treatment planning \cite{berke_doga_basaran_matthews_bai_2022}. This is illustrated in figure \ref{fig:1} where the follow-up scan shows the new lesion delineated in red.

Automatic methods, that is, methods that do not involve manual delineation by experts for longitudinal MS Lesion Segmentation are especially valuable due to the higher costs associated with manual delineation compared to MS Lesion Segmentation using a single time point. However, since new lesions occur infrequently and are typically small, both input and output imbalance are prevalent. In order for automated segmentation methods to be clinically viable, models must address these imbalance issues while demonstrating robustness with limited training data and computational resources. 

The learning process of deep learning models is heavily influenced by the loss function it uses. Therefore, careful design and selection of the loss function is critical. Most automated segmentation methods use the Dice loss   \cite{sudre_li_vercauteren_ourselin_jorge_cardoso_2017} and the Cross-Entropy loss without further consideration. However, this introduces a problem as the limitations of both Dice and Cross-Entropy have been well documented with Dice being insensitive to small lesions and performs poorly in boundary estimation \cite{Zhang_Liu_Li_Wang_2021}. Cross-Entropy and its weighted variants have also been documented to achieve poor performance for highly imbalanced segmentation tasks \cite{loss_odyssey_in_medical_image_segmentation_2021}.  

Input imbalance is the imbalance between the background and the foreground where the foreground is the object to be segmented. This is hardly ever tackled by improving the loss function, and most approaches opt for sampling schemes such as \cite{chawla_bowyer_hall_kegelmeyer_2002, dal2015calibrating}. Region-based loss functions like Dice can handle it to a certain extent due to their formulation but fail to handle it completely and also fail to handle output imbalance effectively \cite{comboLoss_2019}.

The most common approach to handle output imbalance is to introduce an additional parameter to weigh different components. Loss functions like Focal loss \cite{focal_loss_2022}, Tversky loss \cite{salehi2017tversky} and variants based on the two such as Focal-Tversky \cite{focal_tversky_2019} were proposed all with limited success. All of prior functions used an additional parameter to control the trade-off between False Positives (FP) and False Negatives (FN) to try to mitigate output imbalance. Unified Focal loss \cite{Unified_Focal_2022} saw the numerous hyperparmeters as computationally heavy and tried to develop a more efficient loss function by grouping functionally equivalent parameters together. It also used the compound loss strategy by combining the outputs of two loss functions both of which were variants of Focal loss. However, the use of multiple Focal-based losses leads to the model learning inefficiently as both components prioritize the same objective. DicePP loss also used an additional parameter but as an exponent instead of multiplication thus modifying the Dice coefficient, however, the use of a hyperparameter as an exponent leads to vastly slower training due to the number of computations involved when dealing with 3D input thereby making DicePP infeasible to use in practice when resources are scarce \cite{dicePP_2022}. 

Combo loss \cite{comboLoss_2019} however,  used a combination of Dice and a modifed Cross-Entropy to handle both input and output imbalance. The Cross-Entropy term was modified using an additional hyperparameter in order to control the FP-FN trade-off while Dice loss was used for its ability to handle the input class-imbalance problem. However, the performance for longitudinal segmentation is not optimal due to Dice loss weighting FP and FN equally. Further, finer control is needed with longitudinal tasks as the foreground is sparse in most cases and due to the possibility of lesions being present in the first time point that need to be excluded from the segmentation depending on if the task requires segmentation of all lesions in addition to the newly emerging lesions or only the newly emerging lesions. The result of this is that the likelihood of FP tends to be high. 

Thus, we propose HyTver, a novel loss function to handle both input and output imbalance for longitudinal segmentation by using a combination of the Tversky index and Cross-Entropy loss. We apply the weighted Cross-Entropy introduced by \cite{comboLoss_2019} and use the Tversky index, a generalised variant of the Dice coefficient. The Tversky Index generalises Dice to handle FP and FN by introducing a hyperparameter in the denominator. We additionally use another hyperparameter to control the contribution of the two individual losses. 

In medical image segmentation, a common issue is the lack of data due to various problems, including Intellectual Property rights, patient privacy among others \cite{systematic_review_datasets}. Therefore, using pre-trained weights is essential for models to learn well. Hence, we also assess the performance of our loss functions and others under a pre-trained weight setting in addition to comparing HyTver with some of the numerous loss functions used in various works. 

Therefore our contributions are as follows:
\begin{enumerate}
    \item \textbf{A novel loss function to handle input and output imbalance}:
    We propose HyTver Loss, a novel and enhanced formulation based on Combo Loss, designed to improve segmentation accuracy and training robustness in longitudinal settings.
    \item \textbf{Extensive comparison of loss functions}: 
    A comprehensive evaluation of multiple loss functions for the task of longitudinal MS lesion segmentation.

\end{enumerate}

\section{Methodology}
\subsection{HyTver}
Our proposed loss function HyTver is mathematically defined as follows:

{\small
\begin{align*}
        &L_{TI} = 1 - \frac{y\hat{y}}{y\hat{y} + \alpha(1-\hat{y})y + (1-\alpha)\hat{y}(1-y)} \\
        &L_{mCE} = -\frac{1}{N}\sum_i \beta(y_i - \log(p_i)) + (1-\beta)((1-y_i)\log(1-p_i))\\
        &L_{HT} = \gamma L_{mCE} + (1-\gamma) L_{TI}
\end{align*}
}

where $y$ and $\hat{y}$ are the true and predicted segmentation maps, respectively, while $p_i$ and $\hat{p_i}$ are the probabilities and the predicted probabilities of the lesions in the segmentation mask. $\alpha$, $\beta$ and $\gamma$ are tunable hyperparameters.  $L_{mCE}$ on the other hand, refers to a modified Cross-Entropy defined by \cite{comboLoss_2019} while $L_{TI}$ refers to the Tversky Index. $L_{HT}$ refers to the final HyTver loss.

By formulating HyTver in the above manner, we present a simple solution to the imbalance problem as input imbalance is handled by the Tversky Index by measuring the overlap between the ground-truth and the predicted segmentation maps while output imbalance is handled in two places i.e. through $alpha$ in the Tversky Index and $beta$ in the modifed Cross-Entropy. The Tversky Index uses $\alpha$ and $\beta$ to handle the trade-off between False Positives and False Negatives while ensuring the overall calculation of the loss function is not computationally heavy as the parameter is not applied per voxel but on a summarised value representation of the input. While modifying the Cross-Entropy by adding the one-hot encoded ground-truths allows us to penalise FP and FN.

\subsection{Model}
\begin{figure*}[h]
    \centering
    \includegraphics[width=0.9\linewidth, height=0.40\textheight]{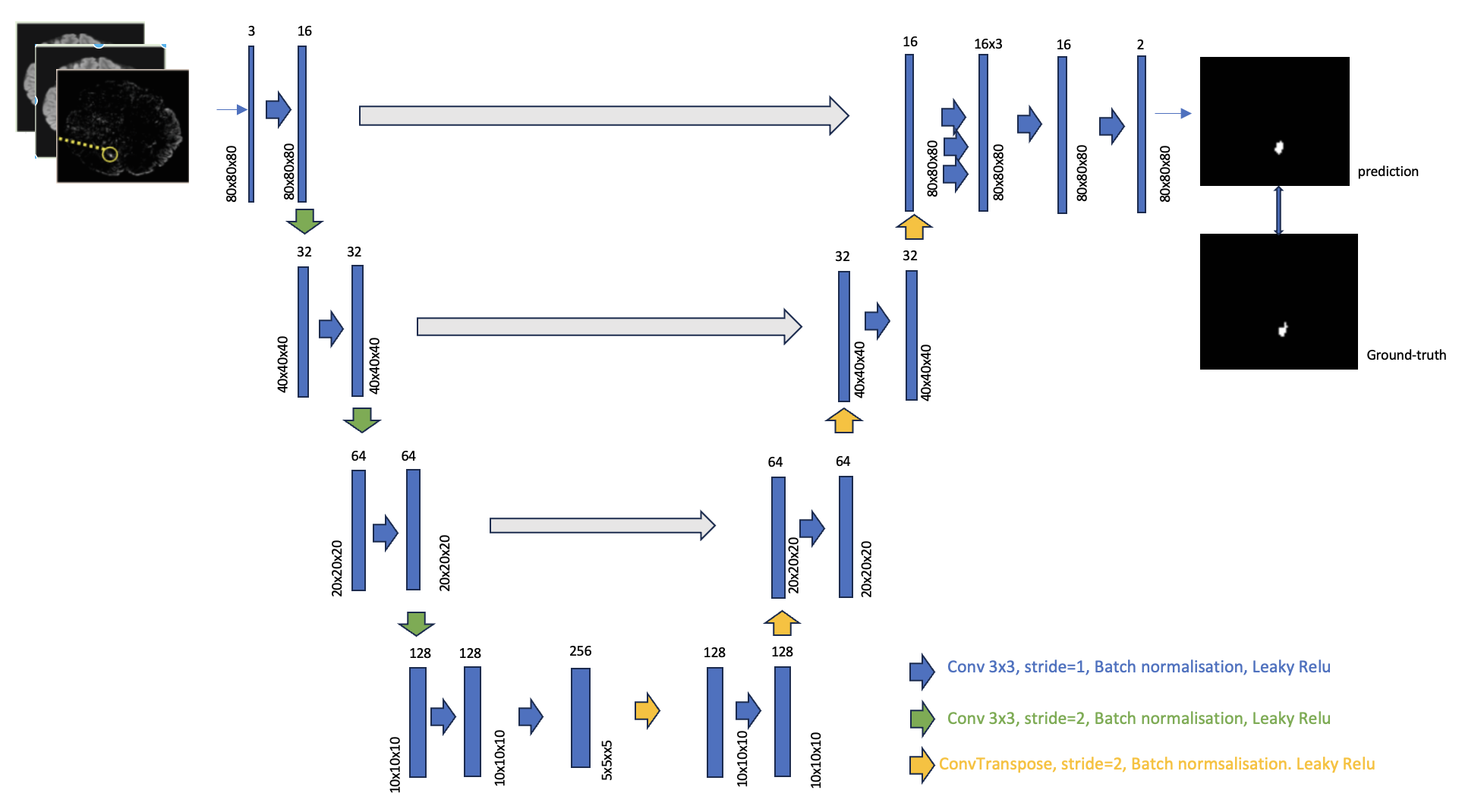}
    \caption{The architecture of the model with the blue boxes representing multichannel feature maps. The arrows represent different operations. The number of channels is depicted on top while the sizes are depicted on the bottom of the boxes.}
    \label{fig:3}
\end{figure*}

We applied a V-Net based model as per figure \ref{fig:3}. It takes the longitudinal difference map as input in addition to the baseline and follow-up images. These longitudinal difference maps are obtained by subtraction between the baseline and the follow-up. All three images are concatenated and then sent as input to the network.

 The convolutional block consists of a 3D convolutional layer followed by a batch normalisation layer  and the leaky ReLU activation function. The downsampling block  is constructed the same way as the convolutional block aside from the number of output filters. The entire upsampling path consists of 5 such blocks.

The convolutional is performed by a convolutional block similar to the one used in the encoder while the upsampling is performed by an upsampling block made up of a transposed convolutional block followed by a batch normalisation layer and the LeakyReLU activation. Skip connections from the corresponding encoder block are also used.

We used LeakyRelu \cite{Maas2013RectifierNI} instead of Relu, as the presence of a small nonzero gradient for input values less than 0 may allow faster convergence\cite{Dubey_Jain_2019}.

\section{Experiments and Results}
\subsection{Setup}
For training, we used the model weights by \cite{wu_wu_shi_picker_chong_cai_2023} and loaded them into the model for transfer learning purposes. We kept all hyper-parameters the same across all runs per experiment. We trained using a batch size of 4, learning rate of 0.001, patch size of 80 and the Adam optimizer. All loss functions were trained for 250 iterations where one iteration. All experiments were performed on a single NVIDIA A5000 GPU.

\subsection{Dataset}
We made use of the Multiple Sclerosis New Lesion Segmentation challenge 2 (MSSEG'2)\footnote{https://portal.fli-iam.irisa.fr/msseg-2/} dataset which is a follow-up to the MSSEG 2016 dataset which wasn't longitudinal\cite{data_2016}. It is a dataset focusing on newly appearing MS lesions with data from two time-steps being provided and contains 365 slices per Magnetic Resonance Imaging(MRI) scan. 

The dataset comprises the MRI data of 100 MS patients. The patients have two scans as demonstrated in figure \ref{fig:1} - the baseline and the follow-up scan - with all scans being of the FLAIR modality. The follow-up scans are taken 1-3 years after the baseline. The scans are taken from 15 different scanners with data from the 3 GE scanners being withheld in order to assess generalisability. The data from 40 patients is available to the public with the rest being kept under wraps as an unseen test set for the challenge itself. Ground-truth was provided for 4 experts as well as a consensus new lesions mask. We only used the latter.

\subsection{Preprocessing}
The organisers of the dataset performed only a rigid brain registration and we further applied a normalisation to a fixed resolution of [0.5, 0.75, 0.75] mm. In addition, a weighted cropping strategy is also used. As for the cropping strategy, if the input sample contained the foreground, one of the foreground voxels were randomly selected as the patch centre and the patch was shifted by a maximum margin of [-10, 10] voxels. If the sample didn't contain foreground, the patch was selected randomly. All patches were of size $80 \times 80 \times 80$. The train-test split was also applied as per \cite{wu_wu_shi_picker_chong_cai_2023} with 32 samples used for training and 8 for validation/testing. Less samples as we didn't have access to their inhouse dataset.

\subsection{Loss Functions}
In addition to our proposed loss function HyTver, we also evaluated other popular loss functions as well. We evaluated four foundational loss functions, their variants and certain popular combinations. The foundational functions being: Dice loss, Focal loss, Binary Cross-Entropy loss and Tversky loss. The rest of the loss functions evaluated were variants of these four proposed by different work. We tried to compare as many as possible in order to provide a complete picture of the performance of various loss functions for longitudinal MS lesion segmentation.

\subsection{Evaluation}
For the purposes of evaluating the loss functions, we used a number of metrics, which included the Dice Coefficient(DC), Jaccard Index(JC), Hausdorff Distance(HD), Average Surface Distance(ASD), Precision(PR) and F1 Score(F1). We mainly focused on the Dc, Hd, and ASD scores as we wanted to evalute our loss functions ability to estimate boundaries well in addition to the overlap performance. From the metrics used, DC and JC measure overlap between the prediction and ground-truth while HD and ASD measure the distances between the prediction and the ground-truth. PR and F1 describe the FP and FN's among the prediction therefore we use these to give a clear picture of segmentation accuracy from a number of different angles.

\subsection{Experiments}

\begin{table}[h]
\caption{Metrics for loss functions}
\label{tab:metrics_all}
\begin{scriptsize}
\begin{tabular}{|l|c|c|c|c|c|c|}
\hline
    \textbf{Loss} & \textbf{DC}~$\uparrow$ & \textbf{JC}~$\uparrow$ & \textbf{HD}~$\downarrow$ & \textbf{ASD}~$\downarrow$ & \textbf{PR}~$\uparrow$ & \textbf{F1}~$\uparrow$ \\
    \hline
    Dice & 0.646 & 0.519 & 45.8 & 11.1 & 0.686 & 0.695 \\
    CE & 0.624 & 0.502 & 37.8 & 16.6 & 0.594 & 0.650 \\
    Focal & 0.633 & 0.513 & 29.1 & 13.7 & 0.626 & 0.687 \\
    Tversky & 0.648 & 0.519 & 45.1 & 10.2 & 0.707 & 0.731 \\
    Symmetric Unified Focal & 0.633 & 0.512 & 29.2 & 13.7 & 0.625 & 0.704 \\
    Asymmetric Unified Focal & 0.633 & 0.512 & 29.2 & 13.7 & 0.625 & 0.704 \\
    Logcosh & 0.633 & 0.512 & 29.2 & 13.7 & 0.625 & 0.704 \\
    DiceCE & 0.649 & 0.520 & 45.8 & 10.6 & 0.686 & 0.702 \\
    FocalDice & 0.646 & 0.517 & 45.1 & 10.2 & 0.696 & 0.700 \\
    FocalTversky & 0.648 & 0.520 & 45.5 & 10.8 & 0.694 & 0.705 \\
    Combo & 0.642 & 0.514 & \textbf{27.2} & 11.4 & 0.688 & 0.724 \\
    Weightedce & 0.625 & 0.503 & 36.8 & 15.5 & 0.590 & 0.663 \\
    HyTver & \textbf{0.659} & \textbf{0.524} & 36.4 & \textbf{6.83} & \textbf{0.724} & \textbf{0.747} \\
    \hline
\end{tabular}
\end{scriptsize}
\end{table}

Table \ref{tab:metrics_all} gives the metric values for all loss functions for the test set.

\section{Discussion}

Longitudinal MS lesion segmentation is prone to both input and output imbalance. Input imbalance occurs due to the small size of the lesions while output imbalance occurs due to FP/FN imbalance. However, for longitudinal segmentation of MS lesion, instead of simply balancing FP and FN, both numbers should be reduced as much as possible. The presence of FN leads to lesions being ignored during segmentation, which has negative consequences in a clinical situation. FP on the other hand, would cause the already existing lesions being segmented as emerging or new lesions which is detrimental for the task of longitudinal MS lesion segmentation.

Referring to table \ref{tab:metrics_all}, the highest Dice score of 0.659 is achieved by HyTver. The incorporation of the Tversky Index enables finer grained control over output imbalance as compared to conventional Dice. This is due to the hyperparameter $\alpha$. This is also true with Tversky loss and Focal Tversky loss performing better than Dice loss by a difference of 0.002 in the Dice score.  Looking at the PR and F1, more proof is provided as HyTver also has the best PR and F1 which indicates the least number of FP and the best balance between PR and Recall respectively. Moreover, looking at the PR and F1 scores of the other functions, Tversky and FocalTversky have better scores compared to Dice and Combo which further prove that incorporating the Tversky Index leads to more accurate and usable results.

Looking at performance from a distance perspective, we need to consider both HD and ASD. This is because HD gives the maximum distance from a point on the predicted boundary to a point on the ground truth while ASD gives the mean of all shortest distances between predicted and true boundaries; therefore, it is less sensitive to outliers compared to HD. This tells us that HyTver has good boundary predictions on average but has a few large outliers as shown by its HD score. Dice, Logcosh, FocalDice and FocalTversky also contain large outliers as their HD scores are the highest among the loss functions however, on average their boundary predictions are reasonable as they have relatively low ASD values. By far the worst performing is CE with high HD and ASD values of 37.8 and 16.6 respectively. The overall best performing by distance is Combo loss with low HD and ASD of 27.2 and 11.4 respectively. This tells us that CE although able to predict the overlap reasonably well, fails at estimating the boundaries of the lesions compared to the other loss functions tested. HyTver on the other hand, is able to have small discrepancies between the boundaries on average with only one major discrepancy as noted by its HD and ASD scores. In fact, HyTver has the lowest ASD score which shows its able to best predict the boundaries on average.

\begin{table}[h!]
\centering
\caption{Comparison of Dice, Precision (PR), and F1 scores across loss functions}
\begin{tabular}{|l|c|c|c|}
\hline
\textbf{Loss Function} & \textbf{Dice} & \textbf{PR} & \textbf{F1} \\
\hline
    asymmetricUnifiedFoc & 0.4270 & 0.5063 & 0.3769 \\
    comboloss & 0.4177 & 0.5035 & 0.4039 \\
    crossEntropy & 0.4620 & 0.5349 & 0.4651 \\
    dice & 0.4011 & 0.4982 & 0.3722 \\
    diceCE & 0.3887 & 0.4904 & \textbf{0.3498} \\
    focal & 0.4575 & 0.5305 & 0.4075 \\
    focalDice & 0.3952 & 0.4878 & 0.3547 \\
    focalTversky & 0.3935 & 0.4970 & 0.3644 \\
    logcosh & 0.4574 & 0.5304 & 0.4110 \\
    symmetricUnifiedFoc & 0.4576 & 0.5306 & 0.4110 \\
    tversky & 0.3877 & 0.4895 & 0.3547 \\
    weightedce & 0.4576 & 0.5344 & 0.4403 \\
    HyTver & \textbf{0.3873} & \textbf{0.4753} & 0.3696 \\
\hline
\end{tabular}
\label{tab:cv}
\end{table}

\begin{figure}
    \centering
    \includegraphics[width=1.0\linewidth]{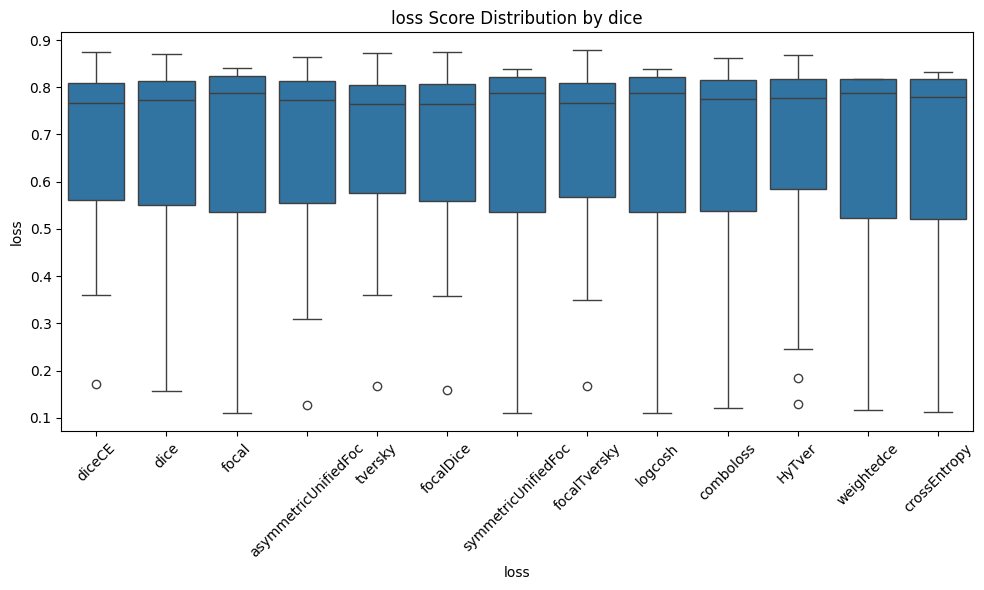}
    \caption{Boxplot for dice scores of each loss function} 
    \label{fig:6}
\end{figure}

\begin{figure}
    \centering
    \includegraphics[width=1.0\linewidth]{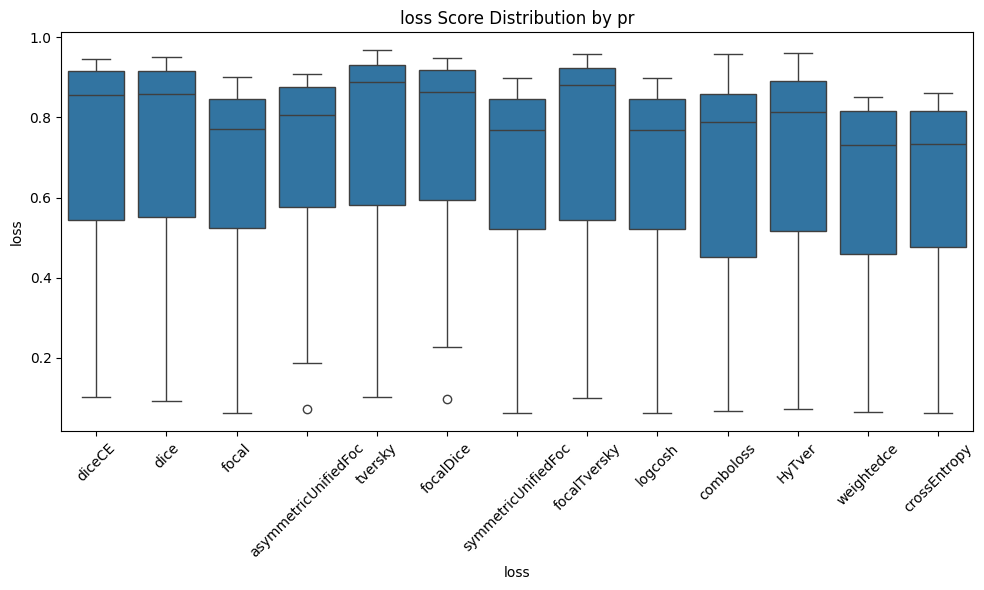}
    \caption{Boxplot for PR scores of each loss function} 
    \label{fig:7}
    
\end{figure}\begin{figure}
    \centering
    \includegraphics[width=1.0\linewidth]{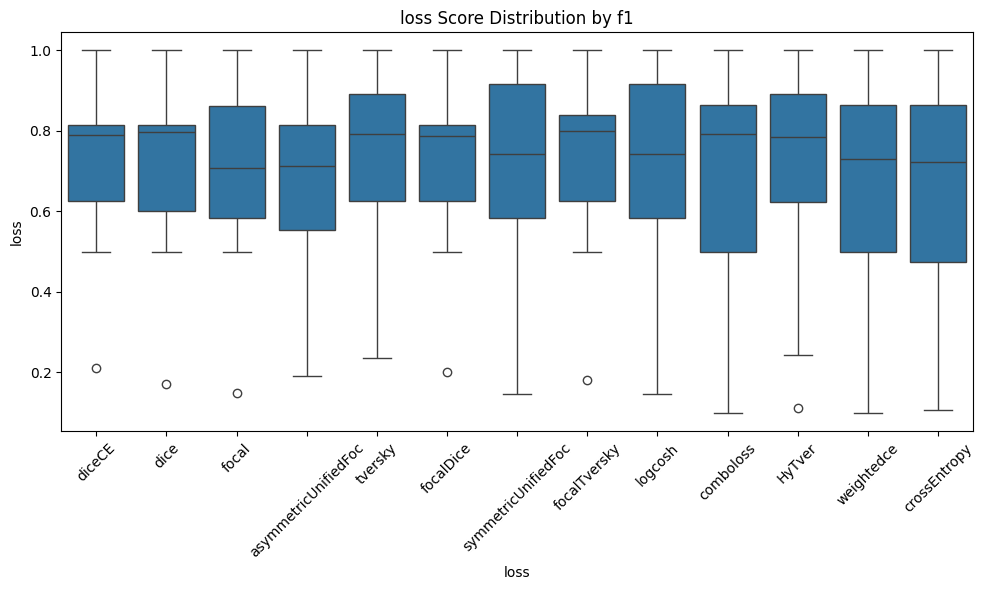}
    \caption{Boxplot for F1 scores of each loss function} 
    \label{fig:8}
\end{figure}

\begin{figure}
\centering
    \begin{subfigure}{.25\textwidth}
        \centering
        \includegraphics[width=\linewidth]{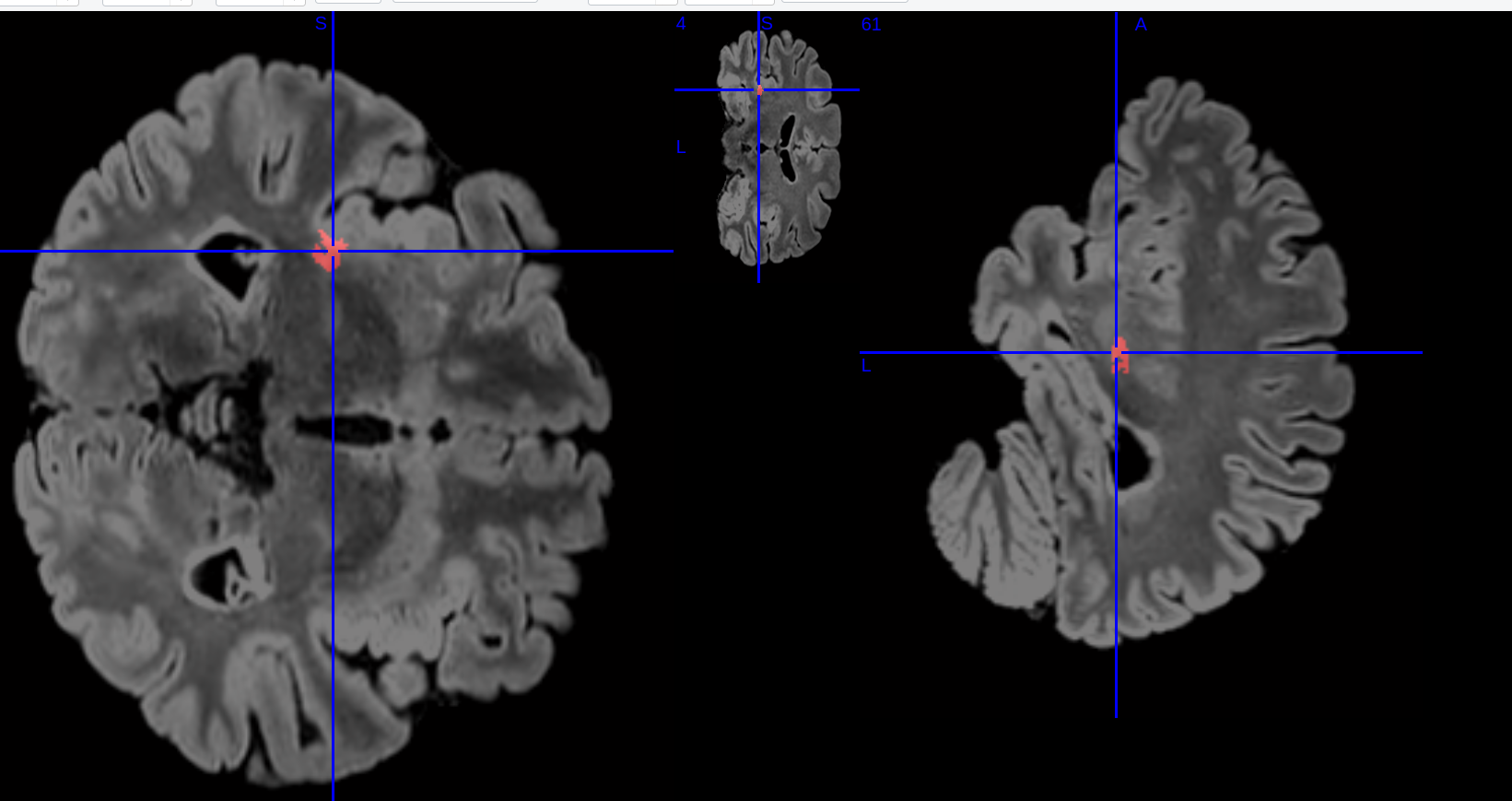}
        \caption{Ground Truth}
        \label{fig:sub1}
    \end{subfigure}%
    \begin{subfigure}{.25\textwidth}
        \centering
        \includegraphics[width=\linewidth]{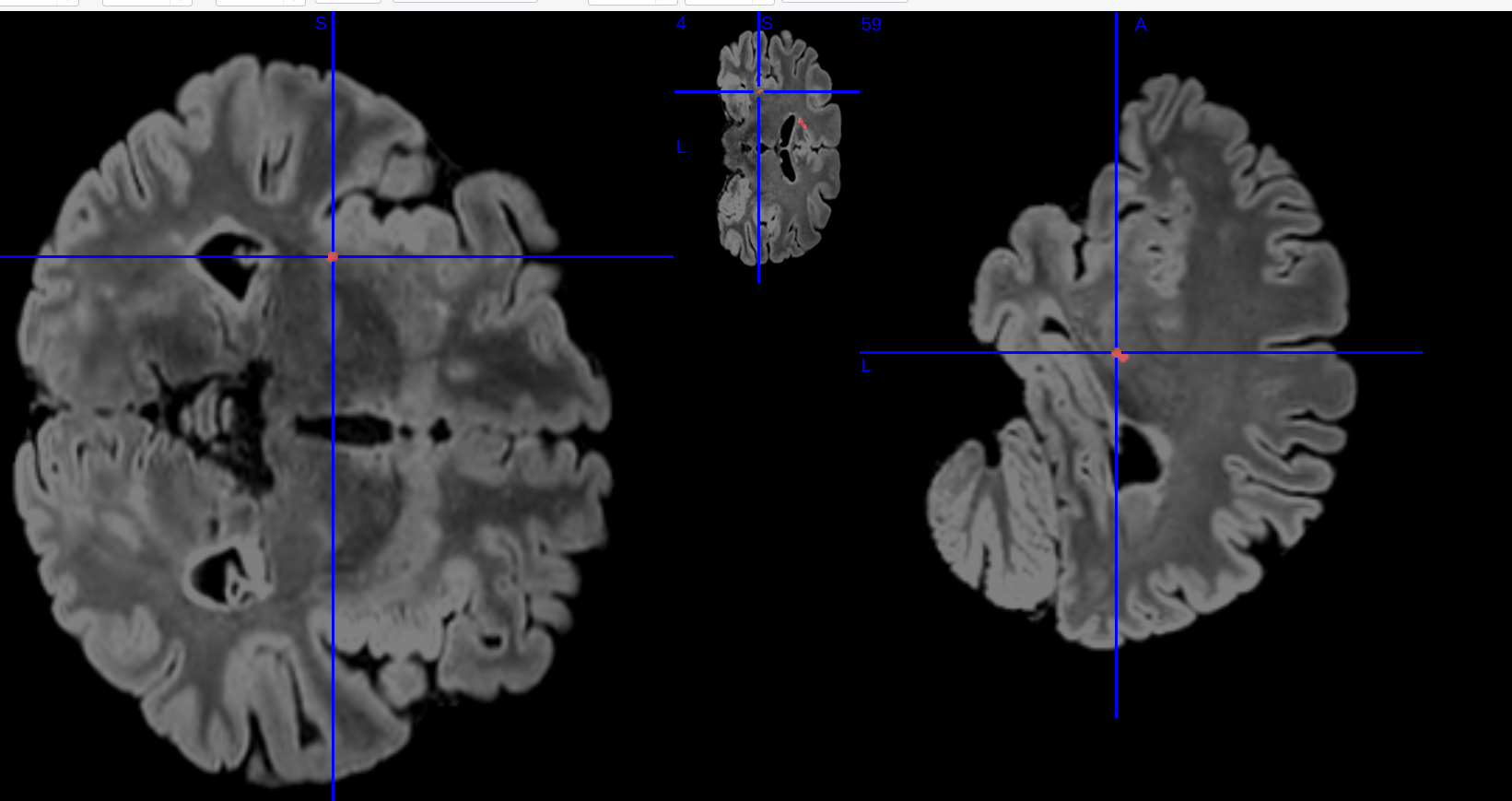}
        \caption{HyTver loss}
        \label{fig:sub2}
    \end{subfigure}
    \medskip
    \begin{subfigure}{.25\textwidth}
        \centering
        \includegraphics[width=\linewidth]{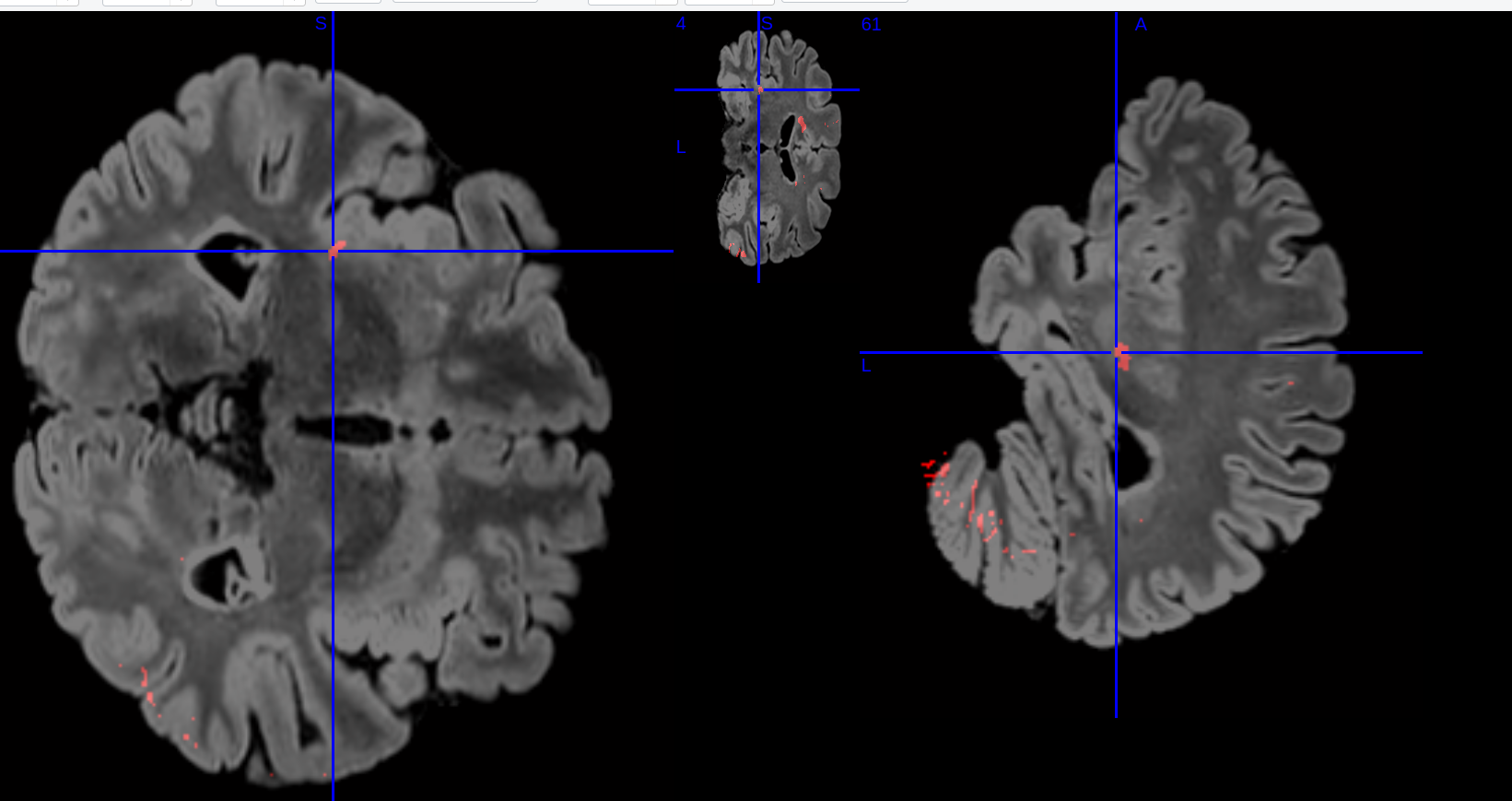}
        \caption{Dice loss}
        \label{fig:sub3}
    \end{subfigure}%
    \begin{subfigure}{.25\textwidth}
        \centering
        \includegraphics[width=\linewidth]{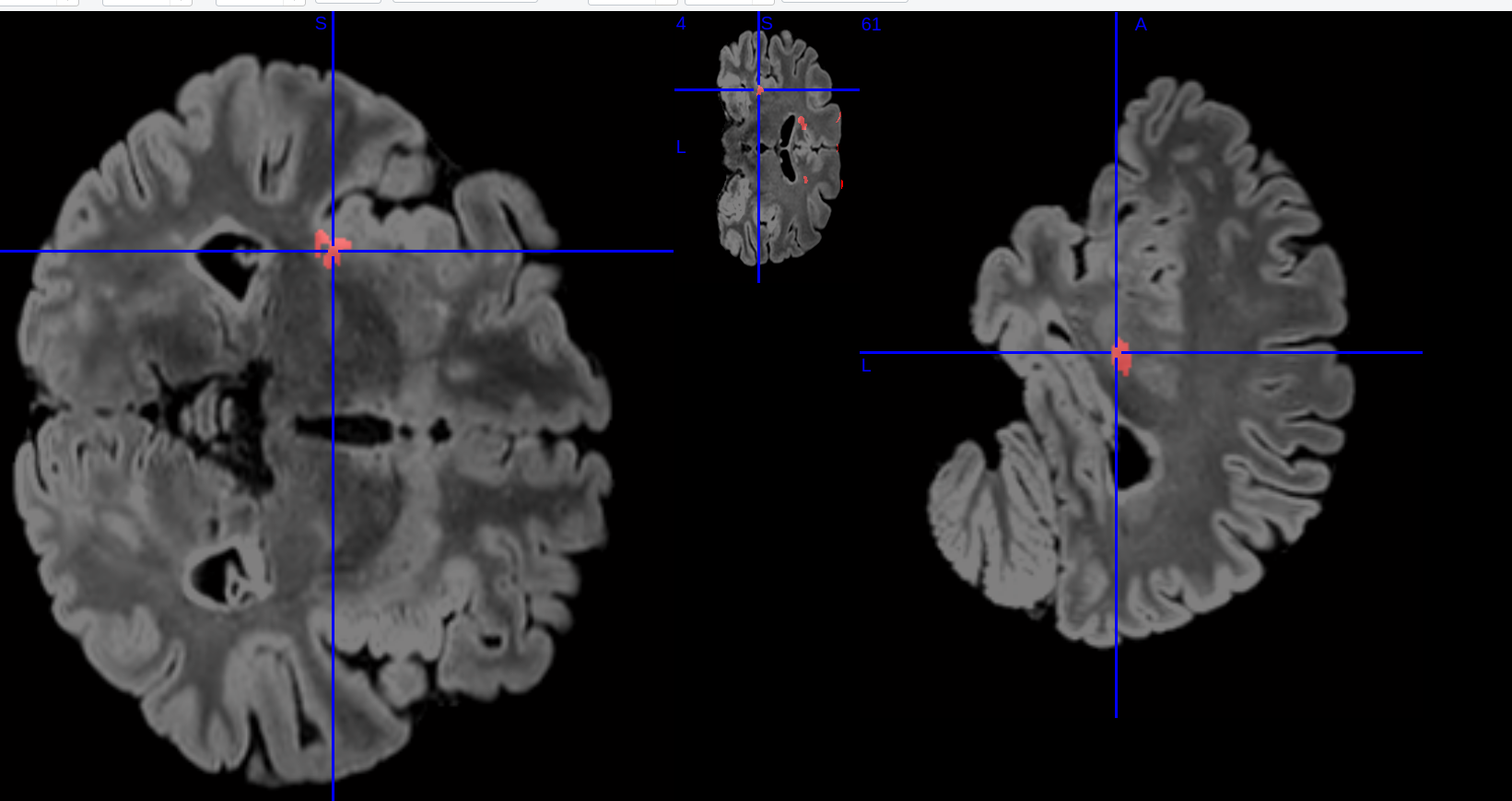}
        \caption{cross entropy loss}
        \label{fig:sub4}
    \end{subfigure}
    \caption{Qualitative comparison of segmentation outputs}
    \label{fig:9}
\end{figure}

Finally, we look at the stability and reliability of the loss functions by looking at their spread, distribution, and CV as given by Figures \ref{fig:6}, \ref{fig:7}, \ref{fig:8} and Table \ref{tab:cv}. HyTver proves to be the most stable considering the CV values of DC and PR given by table \ref{tab:cv} with the CV value of its F1 score also being relatively low compared to other loss functions. This is also shown by the Inter-Quartile-Range (IQR) of the boxplots in figures \ref{fig:6}, \ref{fig:7} and \ref{fig:8} as its IQR is relatively tight in comparison. However, we can see that the IQR of PR is higher than the IQR of F1 for HyTver. This tells us that the balance between PR and Recall (RC) is stable across various samples. In order to avoid FN, a trade-off occurs with PR however, this doesn't prove detrimental as F1 remains stable proving that HyTver is able to achieve a healthy balance between PR and RC consistently. Furthermore, the lack of outliers for PR in HyTver in figure \ref{fig:7} proves that even if it lowers PR to achieve a higher RC, the PR values stay within the acceptable range therefore ensuring the stability of the predictions across patients. The stability of the PR values of HyTver ensures that the loss function does not over-segment the lesions however the F1 values does have an outlier indicating that it might be too cautious and cause the model to under-segment certain cases. This situation is demonstrated in figure \ref{fig:9} where the lesion shape is highly asymmetrical.

Moreover, in comparison to other loss functions, HyTver has a relatively high median value and tighter IQR indicating that HyTver is more stable. This is also supported by table \ref{tab:cv}. Therefore, we can see that HyTver is able to segment the lesions well with good performance in estimating their boundaries. Moreover, it is able to converge well given a set of pre-trained weights. It is also able to maintain a healthy balance of PR and RC and therefore maintain low FP and FN and coupled with low variability ensure that its both accurate and highly stable.

\subsection{Qualitative}
Figure \ref{fig:9} shows a qualitative comparison and we compare HyTver's output with Dice and Cross Entropy loss which are the most commonly used loss functions. We use the sample that got the worst performance output consistently during training. This low score is due to the asymmetrical shape of the lesion which can be difficult to segment. We can see that HyTver has less false positives compared to Dice and Combo loss although it slightly underestimates the size of the leftmost lesion. Dice and Cross Entropy are slightly better at estimating the size of the leftmost lesion, they also come with a much higher number of false positives. Thus HyTver proves to have good segmentation performance while ensuring low false positives and false negatives.

\section{Conclusion}
In this paper, we introduced a new simplified loss function called HyTver that is able to deal with both input and output imbalance. We constructed it using inspiration from Combo loss and using the generalised version of the Dice coefficient which is the Tversky Index. In addition, we also evaluated the performance of different loss functions from different perspectives. HyTver performed the best across most of the metrics including Dice and ASD achieving a Dice score of 0.659. In addition, we also tested the ability of the loss functions to adapt to a task given that the model is initialised with pre-trained weights. In order to analyse this scenario, the CV is used among which HyTver achieves a score for 0.383 for CV for Dice while achieving a competitive performance for the CV for HD and ASD. This proved that HyTver was both performant compared to other popular loss functions and stable. 
For future work, we aim to perform more in-depth testing in order to benchmark for loss functions for the task for medical image segmentation.

\printbibliography

\end{document}